\documentclass{article}

    \PassOptionsToPackage{numbers, compress}{natbib}

\usepackage[preprint]{arxiv}

\usepackage[utf8]{inputenc} 
\usepackage[T1]{fontenc}    
\usepackage{hyperref}       
\usepackage{url}            
\usepackage{booktabs}       
\usepackage{amsfonts}       
\usepackage{nicefrac}       
\usepackage{microtype}      
\usepackage[table,xcdraw]{xcolor}
\usepackage{cleveref}
\usepackage{enumitem}
\usepackage{wrapfig}
\usepackage{booktabs}
\usepackage{graphicx}
\usepackage{rotating}
\usepackage{multirow}
\usepackage{tikz}
\usepackage{tcolorbox}
\usepackage{array}
\usepackage{pgfplots}
\usetikzlibrary{shapes.callouts, shadows}
\usetikzlibrary{shapes.multipart}
\usepackage{listings} 
\usepackage{subcaption}
\usetikzlibrary{arrows.meta, positioning,patterns,arrows}
\usepgfplotslibrary{groupplots}
\usetikzlibrary{backgrounds,calc,bending,backgrounds}
\pgfplotsset{compat=1.16}

\definecolor{cycle1}{RGB}{235,172,35}
\definecolor{cycle2}{RGB}{184,0,88}
\definecolor{cycle3}{RGB}{0,140,249}
\definecolor{cycle4}{RGB}{0,110,0}
\definecolor{cycle5}{RGB}{0,187,173}
\definecolor{cycle6}{RGB}{209,99,230}
\definecolor{cycle7}{RGB}{178,69,2}
\definecolor{cycle8}{RGB}{255,146,135}
\definecolor{cycle9}{RGB}{89,84,214}
\definecolor{cycle10}{RGB}{0,198,248}
\definecolor{cycle11}{RGB}{135,133,0}
\definecolor{cycle12}{RGB}{0,167,108}
\definecolor{cyclegray}{RGB}{189,189,189}

\definecolor{chartreuse}{rgb}{0.75, 1.0, 0.7}
\definecolor{lavender}{rgb}{0.80, 0.80, 0.96}
\usetikzlibrary{graphs,graphs.standard,arrows.meta}

\newcommand{\eg}[0]{\emph{e.g.},}

\newcommand{\syntheticd}[0]{\texttt{ToT-Semantic}}

\newcommand{\arithmeticd}[0]{\texttt{ToT-Arithmetic}}
\newcommand{\ourbenchmark}[0]{\texttt{ToT}}

\usepackage{tikz}

\usepackage{amssymb}
\usepackage{pifont}
\newcommand{\cmark}{\ding{51}}%
\newcommand{\xmark}{\ding{55}}%

\title{Test of Time: A Benchmark for Evaluating LLMs on Temporal Reasoning}

\author{%
  Bahare Fatemi$^1$\thanks{Equal contribution. Corresponding authors: \texttt{baharef@google.com} or \texttt{mehrankazemi@google.com}.}, Mehran Kazemi$^{2*}$, Anton Tsitsulin$^{1}$, Karishma Malkan$^{2}$, Jinyeong Yim$^{3}$, \\ \textbf{John Palowitch$^{2}$, Sungyong Seo$^{3}$, Jonathan Halcrow$^{1}$, and Bryan Perozzi$^{1}$}\\
  $^{1}$Google Research, $^{2}$Google DeepMind, $^{3}$Google \\
}

\begin{document}

\maketitle

\begin{abstract}
Large language models (LLMs) have showcased remarkable reasoning capabilities, yet they remain susceptible to errors, particularly in temporal reasoning tasks involving complex temporal logic. Existing research has explored LLM performance on temporal reasoning using diverse datasets and benchmarks. However, these studies often rely on real-world data that LLMs may have encountered during pre-training or employ anonymization techniques that can inadvertently introduce factual inconsistencies.
In this work, we address these limitations by introducing novel synthetic datasets specifically designed to assess LLM temporal reasoning abilities in various scenarios. The diversity of question types across these datasets enables systematic investigation into the impact of the problem structure, size, question type, fact order, and other factors on LLM performance. Our findings provide valuable insights into the strengths and weaknesses of current LLMs in temporal reasoning tasks. To foster further research in this area, we are open-sourcing the datasets and evaluation framework used in our experiments: \url{https://huggingface.co/datasets/baharef/ToT}.
\end{abstract}




\section{Introduction}
Recent breakthroughs in large language model (LLM) research and applications have been significant~\cite{vaswani2017attention,devlin2018bert,raffel2020exploring,fewshot,touvron2023llama,gpt4,gemini1.0,gemini1.5}. These models, capable of generating new content, have fascinated the AI community, leading to the release of numerous LLMs trained on diverse tasks and data types~\cite{zhao2023survey}.
All of these advancements have led to a growing consensus that LLMs are a pivotal advancement on the path to artificial general intelligence (AGI)~\cite{bubeck2023sparks}. Benchmarking reasoning capabilities in LLMs is therefore a problem of pressing interest to the field  \cite{huang2023reasoning}.

In this work we focus on temporal reasoning, an essential task for intelligent systems across many domains.
Temporal reasoning is focused on understanding reasoning between events in time.
Despite this area's importance, existing temporal reasoning benchmarks do not effectively measure the full scope of temporal reasoning relationships.
Instead they typically rely on question-answering tasks based on Knowledge Graph (KG)-style temporal facts about well-known entities. 

\begin{figure}
    \centering
\newcommand{\myfontsize}{\fontsize{7.5}{8}\selectfont} 

\begin{tikzpicture}[
  user/.style={rectangle callout, callout relative pointer={(-0.5cm,0.5cm)}, fill=lavender, drop shadow, text width=5.5cm, align=left, font=\myfontsize},
  userl/.style={rectangle callout, callout relative pointer={(-0.5cm,0.5cm)}, fill=lavender, drop shadow, text width=6.8cm, align=left, font=\myfontsize},
  chatbot/.style={rectangle callout, callout relative pointer={(0.5cm,0.5cm)}, fill=chartreuse, drop shadow, text width=5.4cm, align=left, font=\myfontsize},
  chatbotl/.style={rectangle callout, callout relative pointer={(0.5cm,0.5cm)}, fill=chartreuse, drop shadow, text width=5.8cm, align=left, font=\myfontsize}
]

\node[userl] (user1) at (0,2) {\textit{\textbf{Prompt}}: Below are the list of head coaches for Chelsea FC. Who was the coach before Pochettino?\\Pochettino: July 2023 to May 2024\\Potter: September 2022 to April 2023\\Tuchel: January 2021 to September 2022\\Lampard: July 2019 to January 2021 and April 2023 to June 2023\\Sarri: July 2018 to June 2019};
\node[chatbotl] (chatbot1) at (0,-0.35) {\textit{\textbf{Model Response}}: The coach before Pochettino was \textbf{Frank Lampard} during his second stint with the club from April 2023 to June 2023.};

\node[text width=6cm] at (0,-1.5) 
    {Grounded prompt (answered correctly)};

\node[user] (user2) at (7.3,2) {\textit{\textbf{Prompt}}: Below are the list of head coaches for a club. Who was the coach before E5?\\E5: July 2023 to May 2024\\E4: September 2022 to April 2023\\E3: January 2021 to September 2022\\E2: July 2019 to January 2021 and April 2023 to June 2023\\E1: July 2018 to June 2019};
\node[chatbot] (chatbot2) at (7.3,-0.35) {\textit{\textbf{Model Response}}: \textbf{E4} was the coach before E5.};

\node[text width=6cm] at (7.5,-1.5) 
    {Abstract prompt (answered incorrectly)};

\draw (4.0,-1.0) -- (4.0,3.2);

\end{tikzpicture}
\vspace{-1.3em}
    \caption{Comparison of the same temporal query using real (left) and anonymized (right) entity names. Gemini Advanced correctly answered the query with real names but failed with anonymized names, suggesting that LLMs might rely on their parametric knowledge to solve temporal tasks.}
    \label{figure:motivation}
\end{figure}
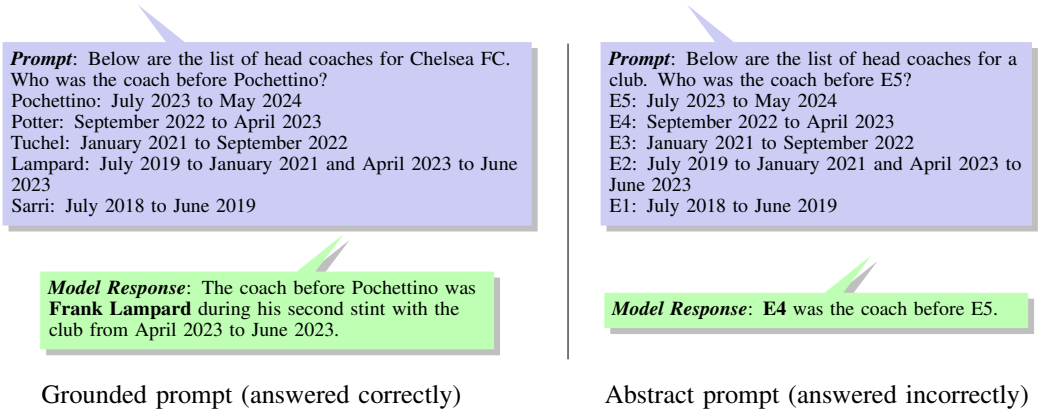

This overemphasis on KG-style temporal facts limits the scope of research and creates several issues.
First, it neglects the diverse 
temporal structure and reasoning tasks found in real-world applications beyond KGs. 
Second, the results on such data often reflect a model's ability to exploit prior knowledge rather than genuine temporal reasoning, making findings less relevant to domains where models lack this knowledge (see \Cref{figure:motivation} as an example.). 
In addition, previous research has shown that shortcuts or heuristics can often answer questions on these datasets without explicit temporal reasoning \cite{chen2022temporal,tan2023towards}. 
Finally, the simple temporal structure of knowledge graphs overlooks the extensive time arithmetic skills required in real-world temporal questions.

\textbf{Our Contributions}: To address these limitations, we develop tasks specifically designed to assess temporal reasoning in a more comprehensive and controlled manner. 
Our benchmark, Test of Time, \ourbenchmark{}, is centered around the observation that temporal reasoning often involves two primary skills: 1) understanding the semantics and logic of time, and 2) the ability to carry out temporal arithmetic.
\ourbenchmark{} has two tasks to cover each essential skill of temporal reasoning, which enable measuring and improving model performances along these two axes independently. \syntheticd{}, a synthetic task, focuses on temporal semantics and logic; it allows for a flexible exploration of diverse graph structures and reasoning task complexity, isolating and evaluating reasoning abilities independent of prior knowledge. \arithmeticd{}, a crowd-sourced task, assesses the ability to perform calculations involving time points and durations. 
Our experimental results with \ourbenchmark{} provide valuable insights into the strengths and weaknesses of current LLMs in temporal reasoning tasks.

\section{Related work}

\textbf{Reasoning.} The ability to draw valid conclusions from explicitly provided knowledge has been a fundamental goal for AI since its early days~\cite{mccarthy1959programs,hewitt1969planner}. In the past few years, several LLM-based techniques have been developed which have advanced the general automated reasoning capabilities of the state-of-the-art models~\cite{wei2022chain,yao2023tree}, or their capabilities in specific directions including mathematical reasoning~\cite{lewkowycz2022solving,ahn2024large}, logical reasoning~\cite{kazemi-etal-2023-lambada,creswell2022selection}, multi-modal reasoning~\cite{wang2024exploring}, commonsense reasoning~\cite{zellers2019hellaswag}, and more. 
Advancing reasoning may explicitly or implicitly translate to improvements in several downstream NLP applications.

\textbf{Temporal reasoning.}
Temporal reasoning has recently gained substantial attention~\cite[e.g.,][]{xiong2024large,beniwal2024remember,jia2024tiq,nylund2023time,hu2023large,gurnee2023language,liu2023grounding,vashishtha2020temporal}. 
Much research focuses on enhancing LLMs' understanding of temporal concepts, primarily through pre-training and fine-tuning strategies to improve their temporal reasoning capabilities~\cite[e.g.,][]{xiong2024large,jia2024tiq,yang2023once,zhou2020temporal,ning2019joint}.
Benchmark creation is another active area, with many benchmarks centered on knowledge graphs~\cite[e.g.,][]{neelam2021sygma,jia2021complex,jia2018tempquestions}.
While TempTabQA~\cite{gupta2023temptabqa} offers crowd-sourced questions based on Wikipedia infoboxes, the process is resource-intensive and prone to issues like LLM overuse by workers.
\citet{xiong2024large} recently proposed TGQA, a data set derived from the YAGO11k knowledge graph \cite{dasgupta2018hyte}. To prevent data leakage, TGQA changes each entity name to a name generated by GPT3.5 that is guaranteed to (i) align with the entity's type and (ii) not be otherwise present in YAGO11k. This strategy has several weaknesses.  
First, it can introduce spurious entity name correlations (LLMs could even potentially guess the original entities due to their adjacent relations).
Second, it can generate factually incorrect or anti-commonsensical claims, for instance if an entity's generated replacement name is a \emph{real} name that is nonetheless not in YAGO11k.
Finally, relying on GPT for copying facts introduces the potential for hallucinations to contaminate the dataset.  



\textbf{Synthetic datasets.} A new trend in probing various LLMs capabilities, especially in the case of reasoning, is through synthetic data that allows for a more systematic evaluation. Previous work has developed synthetic datasets for probing and improving various kinds of reasoning including logical reasoning \cite{tafjord2021proofwriter,kazemi2023boardgameqa,saparov2023testing}and mathematical reasoning \cite{srivastava2024functional,kazemi2023geomverse}.
Most similar to our work, \citet{fatemi2024talk} develop a synthetic probe for measuring the graph-based reasoning abilities of LLMs \cite{sanford2024understanding,perozzi2024let}. 
Our work extends this concept to the case of temporal reasoning with graph-like facts.

\textbf{Present work.} In this work, we introduce \ourbenchmark{}, a novel benchmark for temporal reasoning generated synthetically. Unlike many existing benchmarks that rely on knowledge graphs, \ourbenchmark{} aims to encompass a wider variety of graph structures. Our synthetic generation approach offers precise control over the type of data produced. Importantly, when evaluating LLMs against \ourbenchmark{}, they cannot exploit their latent knowledge for shortcuts; instead, they must genuinely reason with the presented facts. This design promotes a more rigorous assessment of temporal reasoning capabilities in LLMs.

\section{\ourbenchmark{}: A benchmark for evaluating LLMs on temporal reasoning}

We propose that effective temporal reasoning hinges on two distinct skills: understanding the semantics and logic of time, and performing accurate temporal arithmetic. To measure and improve model performance along these independent axes, we create a dedicated task for each skill. By decoupling the evaluation of temporal semantics from arithmetic, we aim to provide a more nuanced analysis of LLM capabilities, pinpointing strengths and weaknesses in each aspect. Experiments on these tasks enable us to independently benchmark LLM performance on both dimensions.

\subsection{\syntheticd{}: A Synthetic Dataset}\label{sec:task-creation}

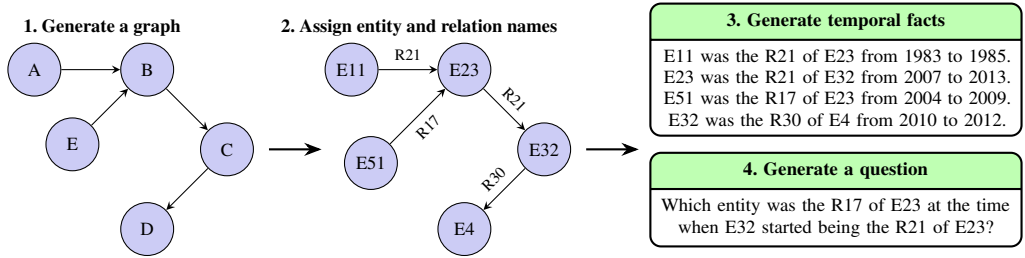
\begin{figure}
    \centering
    \tikzset{
  header node/.style={%
    rectangle split,
    rectangle split parts=2,
    draw,
    rounded corners,  
    thick,            
    rectangle split part fill={#1, white}, 
    text width=4.7cm,
    align=center,
    font=\scriptsize
  }
}

\tikzset{
  large header node/.style={%
    header node=#1, 
    text width=10cm,
    minimum height=4cm  
  }
}

\begin{tikzpicture}[->, >=stealth, node distance=1.5cm and 2cm, xshift=-10cm]

\tikzstyle{graph_node} = [circle, draw, fill=lavender, inner sep=2pt, minimum size=2em, font=\scriptsize]
\tikzstyle{graph_edge} = [left, font=\tiny,sloped,anchor=south,auto=false]

\node[graph_node] (A) {A};
\node[graph_node] (B) [right of=A] {B};
\node[graph_node] (C) [below right of=B] {C};
\node[graph_node] (D) [below left of=C] {D};
\node[graph_node] (E) at ([shift=({225:4 em})]B) {E};

\path (A) edge (B);
\path (B) edge (C);
\path (C) edge (D);
\path (E) edge (B);

\node at ($(B.north)+(-0.6,0.2)$) {\scriptsize\textbf{1.\ Generate a graph}};

\node[graph_node] (A2) [right=3.5cm of A] {E11};
\node[graph_node] (B2) [right of=A2] {E23};
\node[graph_node] (C2) [below right of=B2] {E32};
\node[graph_node] (D2) [below left of=C2] {E4};
\node[graph_node] (E2) at ([shift=({225:5 em})]B2) {E51};

\path (A2) edge node[graph_edge] {R21} (B2);
\path (B2) edge node[graph_edge] {R21} (C2);
\path (C2) edge node[graph_edge] {R30} (D2);
\path (E2) edge node[graph_edge,anchor=north] {R17} (B2);

\node at ($(B2.north)+(-0.6,0.2)$) {\scriptsize\textbf{2.\ Assign entity and relation names}};
\path[-Stealth, thick, black] ($(C.east)+(0.2,0)$) edge ($(C.east)+(0.9,0)$);

\node[header node=chartreuse, right=3.6cm of A2] (box1) {
    \textbf{3.\ Generate temporal facts} 
    \nodepart{second}
E11 was the R21 of E23 from 1983 to 1985.\\
E23 was the R21 of E32 from 2007 to 2013.\\
E51 was the R17 of E23 from 2004 to 2009.\\
E32 was the R30 of E4 from 2010 to 2012.
  };
  
\node[header node=chartreuse, below=of box1, yshift=1.3cm] (box2) {
    \textbf{4.\ Generate a question} 
    \nodepart{second}
    Which entity was the R17 of E23 at the time when E32 started being the R21 of E23?
  };

\path[-Stealth, thick, black] ($(C2.east)+(0.2,0)$) edge ($(C2.east)+(0.9,0)$);
\end{tikzpicture}
    \caption{Steps for creating the \syntheticd{} dataset.}
    \label{figure:time_semantic_dataset}
\end{figure}

The first task we introduce, \syntheticd{}, consists of synthetic problems designed to highlight temporal semantics and logic in reasoning. 
This task is unique because it allows us to freely experiment with a wide range of temporal dependencies and manipulate the difficulty of the reasoning problem. 
This allows us to isolate and analyze the core reasoning capabilities of a LLM, separating them from any reliance on pre-existing parametric knowledge.
To create the \syntheticd{} task, we took the following steps (summarized in \Cref{figure:time_semantic_dataset}):

\textbf{Step 1: Random structure generation.}
We begin by generating random structures that we will then use to create temporal questions.
To ensure we generate a diverse set of random structures for this purpose, we turn to the literature on graph structure generation.
From it, we employ several existing algorithms for generating graphs of varying properties. This includes Erd\H{o}s-R\'enyi~(ER) graphs~\citep{erdds1959random}, scale-free networks~(SFN)~\citep{scalefree}, graphs following the Barabási–Albert~(BA) model~\citep{barabasi}  and stochastic block model~(SBM)~\citep{holland1983stochastic}, as well as star and complete graphs. Each of these graph generation algorithms exhibit different properties and correspond to graphs that appear in different applications. 
For instance, Erdős-Rényi graphs are typically sparse with low average degree, while Barabási-Albert graphs are dense and exhibit power-law degree distributions. We leverage the NetworkX library for generating our random graphs. 
Additionally, we extracted anonymized EgoNets from WikiData~\cite{vrandevcic2014wikidata} by replacing the entity and relation names with generic names. We refer to this structure as \emph{Anonymized Wikidata Extract (AWE)} in our experiments. 
We generate graphs with the number of nodes selected uniformly at random from the [5-30] interval.
More details on the random graph generators used (with visualizations) are available in the Appendix \ref{asec:graph-generators}.

Once we have an initial graph structure, we assign relations to the edges. For each graph, we first decide a number of relation types to be assigned to the edges, and assign each of these relation types to one of one-to-one, one-to-many, many-to-one and many-to-many. Then, for each edge in the graph, we randomly assign between 1 to p (=3 in our experiments) relations types without violating the relation type arity. Then, for each edge $(u, v)$ labeled with a relation $r$, we assign a valid time interval that respects the relation types.

\textbf{Step 2: Question Generation.}
Having generated the random graphs in step 1, we then create questions about those graphs. We consider eight types of questions that are frequently used in day to day life and are common in various benchmarks. \textbf{EventAtTimeT}: asking which entity had some relation R with entity E at some T; \textbf{EventAtWhatTime}: asking at what time a relation R between two entities E1 and E2 started/ended; \textbf{NumberOfEventsInTimeInterval}: asking how many entities had relation R with entity E between T1 to T2; \textbf{BeforeAfter}: asking which entity had relation R with E1 right before/after E2; \textbf{EventAtTimeOfAnotherEvent}: asking when E1 had relation R1 with E2, which entity had relation R2 with E3; \textbf{FirstLast}: asking which entity was the first to have relation R with E; \textbf{RelationDuration}: Asking the k-th time relation R happened between E1 and E2, how long did it last; and \textbf{timeline}: Asking to sort the entities that had relation R with E chronologically.

To create any of the above questions, we keep sampling graphs and fact(s) from the graph until the a proper question of the desired type can be created for that graph and for that fact. For example, to create a \emph{BeforeAfter} question, we keep sampling a graph $G$ and fact $F=(S, R, O, T1, T2)$ until we have a case where there was a unique entity $E$ that had relation $R$ with $O$ right before $[T1, T2]$.

Following the above two steps, we generated $10$ questions per graph generation and per question type. We sorted the facts in five different ways as will be discussed later. This gives as a benchmark with a total of $7 * 8 * 5 * 10 = 2800$ questions, where $7$ is the number of graph generation algorithms, $8$ is the number of question types, $5$ is the number of sorting algorithms, and $10$ is the number of samples we generated.
Example questions of each category type are shown in \Cref{table:synthd-examples}.

\begin{table}[b!]
\caption{Example for each question type in the \syntheticd{} dataset.}\label{table:synthd-examples}
\footnotesize
\centering
\begin{tabular}{p{4.6cm}p{8.5cm}} 
\toprule
\textbf{Question Type} & \textbf{Example} \\ 
\midrule
EventAtTimeT & Find the entity that evidently was the R17 of E69 in year 1932. \\
\midrule
EventAtWhatTime & At what time did E69 start being the R90 of E22?\\
\midrule
NumberOfEventsInTimeInterval &  Find the number of unique entities that were the R82 of E27 between 1952 to 1957. Relations that ended in 1952 or started in 1957 must be counted as well.
\\
\midrule
BeforeAfter &  Immediately before E59, which entity was the R20 of E6?\\
\midrule
EventAtTimeOfAnotherEvent & E94 was the R82 of which entity at the time when E83 started being the R20 of E59?\\
\midrule
FirstLast &  Which entity was the first that was the R35 of E91?\\
\midrule
RelationDuration & When E24 was the R53 of E11 for the 2nd time, for how many years did the relation last? The duration can be computed by subtracting the start time from the end time. \\
\midrule
Timeline & Which entities were the R17 of E69?
\\
\bottomrule
\end{tabular}
\end{table}

\begin{figure}
    \centering
    \resizebox{\textwidth}{!}{



\tikzset{
  header node/.style={%
    rectangle split,
    rectangle split parts=2,
    draw,
    rounded corners,  
    thick,            
    rectangle split part fill={#1, white}, 
    text width=2.8cm,
    align=center,
    font=\scriptsize
  }
}

\tikzset{
  large header node/.style={%
    header node=#1, 
    text width=10cm,
    minimum height=4cm  
  }
}

\begin{tikzpicture}

\node[header node=chartreuse] (box1) {
    \textbf{Seed Set} 
    \nodepart{second}
    We selected a seed set of questions.
  };

\node[header node=lavender, right=of box1] (box2) {
    \textbf{Expand} 
    \nodepart{second}
    The annotators expanded the seed set into a large set of questions. 
  };

\node[header node=chartreuse, right=of box2] (box3) {
    \textbf{Filter} 
    \nodepart{second}
    We filtered knowledge heavy and corner case questions.
  };

\node[header node=chartreuse, right=of box3] (box4) {
    \textbf{Categorize} 
    \nodepart{second}
    We grouped the questions based on the required time arithmetic operations.
  };

\node[large header node=chartreuse, below=of box2, xshift=4.1cm, yshift=0.3cm] (box5) {
    \textbf{Functionalize} 
    \nodepart{second}
    We implemented a functional version of each question, where the input arguments are sampled and final answers are calculated using python libraries.
    \begin{verbatim}
    # EXAMPLE: Add days function
    def add_days(start_time, end_time):
        date = random_date()
        n = random.randint(10,100)
        question = f"If today is {date}, what is the day {n} days from now?"
        answer = current_day + datetime.timedelta(days = n)
        return question, answer
\end{verbatim}
  };

\node[header node=chartreuse, left=of box5] (box6) {
    \textbf{Sample} 
    \nodepart{second}
    We generated a dataset by sampling questions and answers from the codes.
  };

\foreach \i/\j in {1/2,2/3,3/4,4/5,5/6}{
  \draw[->, semithick] (box\i) -- (box\j); 
}
\end{tikzpicture}}
    \caption{Steps for creating the \arithmeticd{} dataset. The green and blue colors represent the operations done by the authors and the annotators respectively.}
    \label{figure:time_arithmetic_dataset}
\end{figure}
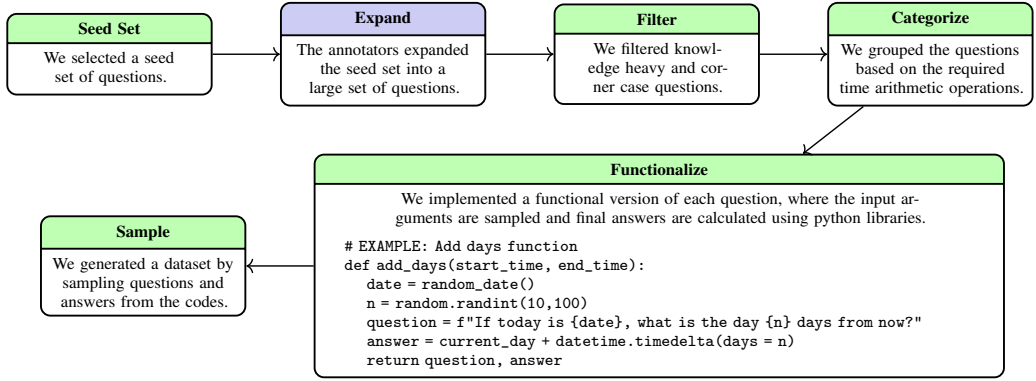

\subsection{\arithmeticd{}: A Temporal Arithmetic Dataset}\label{sec:arithmetic}
Our second task, \arithmeticd{}, shifts from synthetic data to a real-world focus. This task moves beyond understanding the logic and semantics of time and delves into the practical application of mathematical operations within a temporal context.
Through it, we are able to measure a LLM's proficiency in temporal arithmetic and its practical utility in handling time-related computations.

\subsubsection{Task Creation}
To create a large time-arithmetic dataset that covers a wide variety of problems, we took the steps summarized in \Cref{figure:time_arithmetic_dataset}. We explain each step in more detail below.
\begin{itemize}[leftmargin=10pt,topsep=1pt,itemsep=0.2ex,partopsep=1ex,parsep=1ex]
    \item \textbf{Seed Set:} By examining the existing benchmarks and the kind of temporal arithmetic questions that arise in them and through searching the web, we gathered a small set of initial questions. 
    \item \textbf{Expand:} We presented our seed set to 15 annotators who were tasked to propose either new time arithmetic questions that were not in our seed set, or to provide questions corresponding to other scenarios or question templates where one requires to do similar time arithmetic operations to one of the questions in our seed set. We gathered a large list of questions through this process.
    \item \textbf{Filter:} We manually went through all the questions and filtered the ones that were focusing on corner cases, or that required extensive knowledge (\eg{} requiring to memorize the entire calendar).
    \item \textbf{Categorize:} We then grouped the remaining problems into seven categories, shown with examples in \Cref{table:arithmetic-examples}.
    Categories are formed based on the time arithmetic operations required, as follows:
    \textbf{AddSubtract}: adding or subtracting a number (corresponding to days, weeks, minutes, hours, etc.) from a date or time; \textbf{Compare}: comparing a number of dates/times provided in different formats chronologically; \textbf{Duration}: computing the difference between two dates/times; \textbf{Schedule}: finding mutual free spots within multiple blocked times; \textbf{Timezone}: involving dealing with different timezones; \textbf{Trick}: involving questions with slight twists; and \textbf{MultiOp}: involving questions where multiple of the above operations are needed. 
    \item \textbf{Funcionalizing:} Following \cite{srivastava2024functional}, we implemented a functional version of each problem to enable sampling different values for each question and solving based on those values. A functional version of one of our simple problems is provided in \Cref{figure:time_arithmetic_dataset}.
    \item \textbf{Sampling:} We then sampled questions and answers from our functionalized problems. We made the number of samples proportional to the number of different problems that fell under each category. Specifically, we sampled 350 for AddSubtract, 350 for Compare, 200 for Duration, 250 for Schedule, 100 for Timezone, 250 for Trick, and 350 for MultiOp. This resulted in a dataset with 1850 questions in total.
\end{itemize}

\subsection{Quality Check} 
For both tasks, we did multiple rounds of quality checks where we verified 1) whether the generated labels are correct, and 2) whether the question is clear and the provided instructions are sufficient to know in what format the output should be produced. This procedure was done until no more issues could be found in the dataset.

\begin{table}[h!]
\footnotesize
\centering
\caption{Examples for each question type in the \arithmeticd{} dataset.}\label{table:arithmetic-examples}
\begin{tabular}{p{2cm}p{10cm}}
\toprule
\textbf{Category} & \textbf{Example} \\ 
\midrule
AddSubtract & Your driver's license expires on 18 May, 2017. You receive a renewal notice saying it can be renewed 117 days in advance. What's the earliest date you can renew your license? \\
\midrule
Compare &  E42 was discovered in 14 April, 52 BC and E11 was discovered in 05 October, 530 BC. Which was discovered earlier? \\
\midrule
Duration & 	Stella and William were born on 1999-Dec-16 and 2000-Oct-03 respectively. When William was 400 days old, how old was Stella in days?\\
\midrule
Schedule & Lucas is available from 11 to noon and also from 3:30 to 5. Asher is available from 11 to 12:30 and also from 4 to 5. They want to have a 30 minute meeting. The meeting has to start on the hour or half hour. How many possibilities are there for the meeting time?\\
\midrule
Timezone & Flight departs location A at 11:08 (24hr) UTC(+0000). It reaches location B at 07:23:20 PM IST(+0530). What is the total time duration taken to fly?\\
\midrule
Trick & If the date for the day before tomorrow in yyyy-mm-dd format is 2016-01-20, what is the date 27 days from now in the same format?\\
\midrule
MultiOp & Alex solves 2 puzzles in 4 hours, 50 minutes, and 22 seconds. What is the time taken by them to solve 6 puzzles, at the same pace.\\
\bottomrule
\end{tabular}
\end{table}

\section{Experiments and Results}
In this study, we evaluate the performance of three frontier large language models (LLMs) on our benchmark. The models evaluated include  Claude-3-Sonnet~\cite{anthropic2024claude3haiku}, GPT-4~\cite{gpt4}, and Gemini 1.5 Pro~\cite{gemini1.5}. 
As the \syntheticd{} task requires a large context size,  for this benchmark we used the GPT4-Turbo because of its higher context limit; For \arithmeticd{} we used GPT4 due to its higher quality.

In our experiments, we aim to answer the following questions:

\begin{wraptable}[11]{r}{0.5\textwidth}
\vspace*{-4.0ex}
\caption{LLM accuracy on temporal reasoning tasks by graph structure.}\label{table:graph-structure}
\resizebox{0.5\textwidth}{!}{%
\setlength{\tabcolsep}{3pt}
\begin{tabular}{lccc}
\toprule
Graph  & Claude-3-Sonnet & GPT-4 & Gemini 1.5 Pro \\
\midrule
BA                & 48.50      & \textbf{63.25}         & 62.75            \\
Complete          & 34.00      & 40.25         & \textbf{52.50}               \\
ER                & 42.25      & 48.75         & \textbf{60.50}              \\
SBM               & 42.00      & 50.75         & \textbf{57.75}             \\
SFN               & 58.75      & 75.25         & \textbf{75.75}         \\
Star              & 59.50      & \textbf{80.25}         & 74.25        \\
AWE               & {68.75}      & \textbf{92.00}         & {87.50}               \\
\midrule
\textbf{Average Rank}  & 3.00      & 1.57   & \textbf{1.43}                 \\
\bottomrule
\end{tabular}
}
\end{wraptable}

\begin{itemize}[left=0pt]
    \item \textbf{RQ1:} What is the effect of the temporal structure on the LLM performance?
    \item \textbf{RQ2:} What kind of temporal questions are easier/harder for LLMs to answer?
    \item \textbf{RQ3:} How important is the order of the facts in the model prompt and what is the best way of ordering the facts?
    \item \textbf{RQ4:} How well do the frontier models perform on two aspects of temporal reasoning: semantics and arithmetic?    
\end{itemize}

\subsection{Investigating the impact of temporal structure on LLM temporal reasoning}\label{sec:graph-structure}

In different applications where temporal reasoning arises, the structure of the facts can be different. Some tasks may provide all the information about an entity (corresponding to a star graph) and then asks questions about it, whereas in some applications such as social networks the structure of the facts may follow a power-law distribution. It is natural to question whether the inherent temporal structure of a problem might influence an LLM's ability to reason over its data. Drawing inspiration from recent work analyzing graph neural networks~\citep{palowitch2022graphworld,tsitsulin2022synthetic,yasir2023examining,fatemi2024talk}, this section aims to quantify how different temporal dependencies affects an LLM's temporal reasoning capabilities using graph generators to create many different kinds of temporal structure.
Our findings demonstrate that graph structure can significantly impact an LLM's reasoning performance. 

The graph structure of the temporal relationships significantly affects LLM performance, as demonstrated in Table \ref{table:graph-structure}. Notably, GPT-4 accuracy more than doubled between complete graphs ($40.25\%$) and AWE graphs ($92.00\%$).
GPT-4 accuracy varied drastically across graph types, from $20.33\%$ for complete graphs to $90.83\%$ for AWE graphs. This highlights a critical gap in temporal reasoning research, which has largely overlooked the diverse graph structures and reasoning tasks found in real-world applications, instead focusing primarily on specific knowledge graphs (like YAGO11k). This may explain the superior performance of LLMs on AWE graphs in our experiments.


\subsubsection{Influence of graph size on LLM performance}

\begin{wraptable}[12]{r}{0.25\textwidth}
\vspace*{-2.75ex}
\caption{Average number of nodes and edges by graph structure.}\label{table:graph-structure-nnodes-nnedges}
\resizebox{0.25\textwidth}{!}{%
\setlength{\tabcolsep}{3pt}
\begin{tabular}{lcc}
\toprule
Graph  & $\#$nodes & $\#$edges\\
\midrule
BA                &       17.41 & 144.07        \\
Complete          &   17.25 & 619.86         \\
ER                &    16.18 & 316.4        \\
SBM             &    17.51 & 368.15       \\
SFN              &  17.52 & 53.46     \\
Star              &      16.16 & 34.12    \\
AWE               &   18.99 & 25.41            \\
\midrule
\textbf{Average}  &    17.29 & 223.07               \\
\bottomrule
\end{tabular}
}
\end{wraptable}

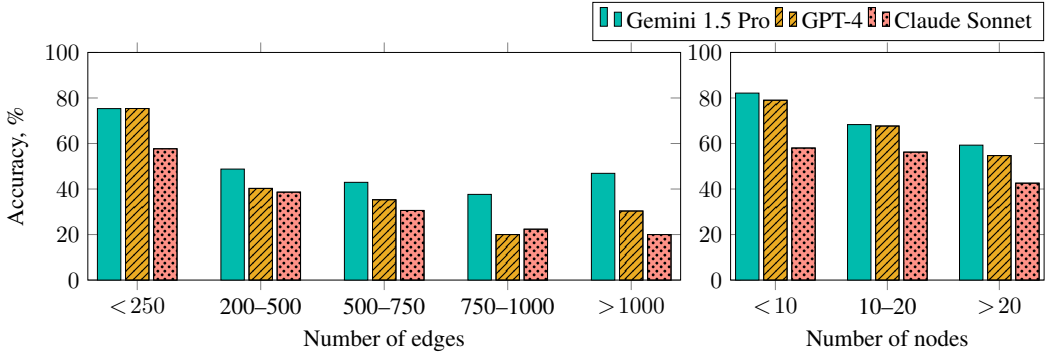
\begin{figure*}[t]
\centering
\resizebox{\linewidth}{!}{\begin{tikzpicture}
\begin{groupplot}[group style={
                      group name=myplot,
                      group size= 2 by 1, horizontal sep=0.75cm},height=5cm,ymax=100,xtick=data,ymin=0,ymax=100,ybar,legend cell align=left,ybar]
\nextgroupplot[
 	ylabel={Accuracy, \%},
 	xlabel={Number of edges},
 	symbolic x coords={$<\!250$, 200--500, 500--750, 750--1000, $>\!1000$},
 	width=0.75\linewidth,
 	]
\addplot[fill=cycle5] coordinates {
    ($<\!250$,75.33169533169534)
    (200--500,48.76712328767123)
    (500--750,42.94117647058823)
    (750--1000,37.64705882352941)
    ($>\!1000$,46.89655172413793)
};
\addplot[fill=cycle1,postaction={pattern=north east lines}] coordinates {
    ($<\!250$,75.38083538083538)
    (200--500,40.273972602739725)
    (500--750,35.294117647058826)
    (750--1000,20.0)
    ($>\!1000$,30.344827586206897)
};
\addplot[fill=cycle8,postaction={pattern=crosshatch dots}] coordinates {
    ($<\!250$,57.690417690417696)
    (200--500,38.63013698630137)
    (500--750,30.58823529411765)
    (750--1000,22.35294117647059)
    ($>\!1000$,20.0)
};
\nextgroupplot[
 	xlabel={Number of nodes},
 	symbolic x coords={$<\!10$, 10--20, $>\!20$},
 	width=.45\linewidth,
 	enlarge x limits=0.2,
 	legend style={at={(1,1.05)},anchor=south east},
 	legend columns=3,
 	]
\addplot[fill=cycle5] coordinates {
    ($<\!10$,82.14876033057851)
    (10--20,68.3157894736842)
    ($>\!20$,59.27710843373494)
};
\addplot[fill=cycle1,postaction={pattern=north east lines}] coordinates {
    ($<\!10$,79.00826446280992)
    (10--20,67.6842105263158)
    ($>\!20$,54.69879518072289)
};
\addplot[fill=cycle8,postaction={pattern=crosshatch dots}] coordinates {
    ($<\!10$,58.01652892561984)
    (10--20,56.21052631578948)
    ($>\!20$,42.570281124497996)
};
\legend{Gemini 1.5 Pro,GPT-4,Claude Sonnet}
\end{groupplot}
\end{tikzpicture}}
\caption{Accuracy of models for different number of edges and nodes.}
\label{fig:nnodes_nnedges}
\end{figure*}

A key question is whether the size of a graph, measured by the number of edges~(facts) and nodes~(entities) respectively, impact LLM performance. As illustrated in \Cref{fig:nnodes_nnedges}, increasing either the number of edges or nodes in the \syntheticd{} dataset leads to a decrease in LLM performance.

This raises the question of whether the graph structure's impact observed in \Cref{sec:graph-structure} is merely a consequence of varying graph sizes. To address this, we present the average number of nodes and edges for each graph structure in \Cref{table:graph-structure-nnodes-nnedges}. While the average number of nodes does not appear to consistently influence LLM performance across structures, the number of edges does show some correlation. However, there are exceptions. For instance, SFN graphs, despite having far fewer edges on average than Star graphs, exhibit slightly better average performance in \Cref{table:graph-structure}. This indicates that both the number of edges and the specific structure of the graph play a significant role in determining LLM performance.

\subsection{Effects of temporal question type on LLM temporal reasoning}

\begin{wraptable}[12]{h}{0.7\textwidth}
\vspace*{-3.7ex}
\caption{LLM accuracy on temporal reasoning by question category.}\label{table:temporal-task}
\resizebox{0.7\textwidth}{!}{%
\setlength{\tabcolsep}{3pt}
\begin{tabular}{lccc}
\toprule
Temporal Question Type & Claude-3-Sonnet & GPT-4 & Gemini 1.5 Pro\\
\midrule
EventAtTimeT                      &   47.14    &   65.43    & \textbf{72.29} \\
EventAtWhatTime                   &   {90.29}    &   {89.43}    & \textbf{93.14} \\
\midrule
NumberOfEventsInTimeInterval  &   29.71    &   \textbf{61.43}    & 59.14 \\
BeforeAfter                           &   53.14      &   \textbf{55.43}    & 52.86 \\
EventAtTimeOfAnotherEvent&    50.00   &   67.14    & \textbf{71.43} \\
FirstLast                             &  \textbf{68.57}     &   67.71    & \textbf{68.57} \\
RelationDuration                      &  41.43     &    {80.00}   & \textbf{84.57} \\
Timeline                                &  24.00     &    28.29   & \textbf{36.29} \\
\midrule
\textbf{Average Rank}                                 &  2.56     &   2.00    & \textbf{1.44} \\
\bottomrule
\end{tabular}
}
\end{wraptable}

In this experiment, we systematically investigate the impact of different temporal tasks on the reasoning ability of LLMs. We quantify this impact by evaluating model performance across a variety of tasks, as summarized in~\Cref{table:temporal-task}.

\textbf{Task type and reasoning requirements.}
A key question in our investigation is whether the type of temporal task and the associated reasoning requirements influence LLM performance. The \syntheticd{} dataset includes questions of varying difficulty levels, which can be categorized based on the nature of the reasoning involved: \textbf{Single-fact solutions:} Questions EventAtTimeT and EventAtWhatTime require retrieving one single fact and answering the question directly based on the retrieved fact. \textbf{Multi-fact solutions:} The remaining questions require retrieving multiple facts and performing operations (\eg{} counting, sorting) to extract the relevant information and formulate an answer.

LLMs consistently demonstrate superior performance on tasks requiring the retrieval of a single fact compared to those necessitating the integration of multiple facts. This performance gap can be attributed to the increased cognitive demands associated with multi-fact tasks. While single-fact questions primarily rely on the identification and extraction of relevant information, multi-fact questions demand a deeper comprehension and synthesis of retrieved information.

\textbf{Performance variations within question types.}
Even within zero-order reasoning tasks, LLMs demonstrate varying levels of proficiency. For example, EventAtTimeT and EventAtWhatTime are structurally similar, yet LLMs tend to excel at the latter.
We hypothesize that this performance difference may be attributed to the fact that EventAtTimeT requires a simple time arithmetic operation to recognize that a timestamp $T$ falls within a time interval $[T1, T2]$, whereas EventAtWhatTime does not require any time arithmetic operation.

\begin{wraptable}{r}{0.5\textwidth}
\vspace*{-2.5ex}
\caption{LLM precision and recall on timeline questions.
}\label{table:timeline-results}
\resizebox{0.5\textwidth}{!}{%
\setlength{\tabcolsep}{3pt}
\begin{tabular}{l|cc|cc}
\toprule
 & \multicolumn{2}{c}{All} & \multicolumn{2}{c}{Complete} \\
Graph structure & Precision & Recall & Precision & Recall\\
\midrule
Claude-3-Sonnet & 0.73  &  0.75  &  0.56 & 0.54\\
GPT-4          &   0.60    &  0.56 &   0.36   &  0.23\\
Gemini 1.5 Pro   &    \textbf{0.81}   &  \textbf{0.83} &   \textbf{0.82}   &  \textbf{0.65} \\
\bottomrule
\end{tabular}
}
\end{wraptable}

\textbf{Analysis on Timeline questions.}
An analysis of timeline questions in \Cref{table:temporal-task} reveals that they pose the greatest challenge across all tasks. The answer to these questions, typically structured as ``Sort the entities that were the R17 of E69 chronologically?'', is often a list of entities. In the \syntheticd{} dataset, every timeline question has more than one entity in its label. 
To further evaluate performance on these multi-answer questions, we calculated the average precision and recall for each model in \Cref{table:timeline-results}. We report the results once averaged over all graph structures and once only for complete graphs (the most challenging graph structure). Gemini 1.5 Pro demonstrates superior precision and recall, aligning with its higher accuracy observed in \Cref{table:temporal-task}. Notably, GPT-4, despite having higher accuracy than Claude-3-Sonnet on timeline questions, exhibits the lowest precision and recall. This suggests that GPT-4 frequently outputs fewer entities than are present in the true answers ($50\%$ of the times), leading to missed correct entities (lower recall) and a higher proportion of false positives among its predictions (lower precision).

Since complete graphs pose the greatest difficulty among all graph structures (\Cref{table:graph-structure}), we provide a separate analysis of average precision and recall for these graphs in the final two columns of \Cref{table:timeline-results}. Notably, Claude-3-Sonnet and GPT-4 experienced declines in both precision and recall on complete graphs, whereas Gemini was primarily impacted in terms of recall.

\textbf{Analysis on NumberOfEventsInTimeInterval questions.} When evaluating models on NumberOfEventsInTimeInterval questions, a clear pattern emerged. Claude-3-Sonnet frequently overestimated the correct count.  In contrast, GPT-4 and Gemini 1.5 Pro demonstrated higher accuracy, although minor errors were observed: Gemini 1.5 Pro tended towards slight overestimation, and GPT-4 towards slight underestimation.

\begin{wraptable}{r}{0.6\textwidth}
\vspace*{-3.0ex}
\caption{LLM accuracy on temporal reasoning tasks as a function of the order of the facts.}\label{table:order-facts}
\resizebox{0.6\textwidth}{!}{%
\setlength{\tabcolsep}{3pt}
\begin{tabular}{lccc}
\toprule
Order of facts & Claude-3-Sonnet  & GPT-4 & Gemini 1.5 Pro\\
\midrule
Shuffle                     &   45.71    &  60.71     & \textbf{63.04} \\
RelationAndStartTime  &   54.29    &  65.36     & \textbf{68.57} \\
StartTimeAndRelation	&   47.68    &  60.54     & \textbf{64.64} \\
StartTimeAndTarget    &   49.11    &  61.61     & \textbf{65.18} \\
TargetAndStartTime    &   {73.57}    &  {62.60}     & \textbf{75.00} \\
\midrule
\textbf{Average Rank}                     &   2.80    &  2.20    & \textbf{1.00} \\
\bottomrule
\end{tabular}
}
\end{wraptable}

\subsection{Impact of temporal fact order on LLM performance}

A noteworthy question arises regarding the potential influence of fact order on LLM performance in temporal reasoning tasks. For investigating this, we conducted experiments on \syntheticd{} dataset.
We sorted the facts using different methods:
\textbf{Shuffle:} randomizing the order of facts; \textbf{RelationAndStartTime:} prioritizing facts based on their relation name, then by start time; \textbf{StartTimeAndRelation:} prioritizing facts based on start time, then by relation name;
\textbf{StartTimeAndTarget:} prioritizing facts based on start time, then by the target entity;
\textbf{TargetAndStartTime:} Prioritizing facts based on the target entity, then by start time.

Ideally, LLMs should exhibit robustness to the order in which facts are presented, given the independent nature of each fact. However, as shown in~\Cref{table:order-facts}, our observations reveal a significant impact of fact order on LLM performance. Notably, performance is consistently lowest when facts are presented in a shuffled order and consistently highest when facts are sorted based on the target entity and start time (TargetAndStartTime). This finding offers valuable practical insights into how facts should be structured when temporal reasoning is a key component of the LLM task. By organizing facts in a manner that aligns with the temporal flow of the narrative or task, we can potentially enhance LLM performance and ensure more accurate and reliable results. While previous work has shown that ordering premises in the correct order of chain-of-thought solution improves LLM's logical reasoning \cite{chen2024premise,saparov2022language}, our results extend that to general-purpose temporal orderings (independent of the chain-of-thought).

\begin{wraptable}{r}{0.5\textwidth}
\vspace*{-2.5ex}
\caption{LLM accuracy on the \arithmeticd{} dataset by question type.}\label{table:arithmetic-results}
\resizebox{0.5\textwidth}{!}{%
\setlength{\tabcolsep}{3pt}
\begin{tabular}{lccc}
\toprule
Category & Claude-3-Sonnet  & GPT-4 & Gemini 1.5 Pro\\
\midrule
AddSubtract           &     58.57      &     \textbf{76.28}      & 71.14 \\
Compare                 &     39.14      &     \textbf{63.14}      & 55.43 \\
Duration                &     15.00      &     \textbf{16.00}      & 13.50 \\
Schedule                &     29.60      &     \textbf{43.60}      & 40.00 \\
Timezone                &      {74.00}     &     {88.00}      & \textbf{90.00} \\
Trick                   &    40.40       &     \textbf{45.60}      & 41.20 \\
MultiOp               &     26.57      &     46.86      & \textbf{62.57} \\
\midrule
Average Rank                 &    2.86       &     \textbf{1.29}      & 1.86\\
\bottomrule
\end{tabular}
}
\end{wraptable}

\subsection{Temporal semantics vs temporal arithmetic}
This study examined the performance of temporal arithmetic capabilities in LLMs using the \arithmeticd{} dataset. 
Results, as shown in~\Cref{table:arithmetic-results}, indicate that the models consistently excelled in Timezone questions, while struggling the most with Duration questions. This superior performance in Timezone questions could be attributed to the abundance of information about various timezones available online, compared to other question types. Scheduling and Trick questions also proved challenging for LLMs, likely due to their creative nature and requirement for deeper reasoning. In contrast, AddSubtract results were relatively strong, potentially reflecting LLMs' optimization for mathematical reasoning and their ability to apply that knowledge to temporal reasoning tasks.

Compared with the semantic questions results from \Cref{table:temporal-task}, we see that there is a substantial difference in average rank results, with GPT-4 doing better on arithmetic and Gemini 1.5 Pro performing better in the reasoning category.

\textbf{Analysis on Duration questions.} Analysis of Duration questions in the \arithmeticd{} dataset revealed them to be the most challenging for the evaluated models. Notably, the most common error among incorrect answers was a deviation of precisely one day from the ground truth label. 
Specifically, when GPT-4 or Gemini 1.5 Pro erred on Duration questions, approximately $21\%$ and $25\%$ of its responses were within one day of the ground truth, respectively.
This suggests that LLMs can approximate the correct calculation but often stumble in the final steps, highlighting a gap in their ability to execute complex arithmetic with precision.

\textbf{Common failure: direction.}
One frequent error in \arithmeticd{} occurs when determining the number of months between two dates. For example, from February 11th, 2002, to October 11th, 2002, the correct duration is eight months, but the model sometimes incorrectly calculates it as four months. This issue is particularly noticeable in questions that involve going back in time, such as: ``Sam's birthdate is October 11th, 1996. Today is February 25th, 2002. Calculate Sam's age in days.''

\textbf{Common failure: leap year calculation.}
Another frequent error in \arithmeticd{} arises when determining the number of days between two dates that span multiple years. Incorrectly accounting for leap years, which have an extra day (February 29th), often leads to inaccurate results. 

\section{Conclusion}
In conclusion, we introdwuce Test of Time (\ourbenchmark{}), a novel benchmark designed to assess LLMs' temporal reasoning abilities in a more comprehensive and controlled manner than existing work.
Our two-pronged approach, encompassing both semantic and arithmetic tasks, enables a nuanced evaluation of temporal reasoning. Through extensive experiments with \ourbenchmark{}, we have gained valuable insights into the strengths and weaknesses of current LLMs in these critical aspects of temporal reasoning. By open-sourcing our datasets and evaluation framework, we hope to stimulate further research and development in this field, ultimately contributing to the advancement of LLM capabilities in complex reasoning tasks.

\section{Acknowledgement}
We express our sincere gratitude to Don Metzler, Radu Soricut, Anastasios Kementsietsidis, Silvio Lattanzi, and Vahab Mirrokni for their great feedback.
We also extend our thanks to all participants who generously provided data for the time arithmetic dataset.

\bibliography{mybib.bib}
\bibliographystyle{plainnat}




\newpage
\appendix

\section{Detailed comparison of \ourbenchmark{} with existing benchmarks.}\label{asec:benchmark-comparison}
\begin{table}[h!]
\caption{Comparison of \ourbenchmark{} against related benchmarks}.\label{table:benchmark-comparison}
\footnotesize
\centering
\begin{tabular}{c|c|c|c|c|c|c}
\toprule
Benchmark & Semantics & Arithmetic & Real-World & Synthetic & Hermetic & Implicit \\
\midrule
StreamingQA \citep{liska2022streamingqa} & \cmark & \xmark & \cmark & \xmark & \xmark & \xmark \\\newline
TimeSensitiveQA \citep{chen2021dataset} & \cmark & \xmark & \cmark & \xmark & \xmark & \xmark \\\newline
TempLama \citep{dhingra2022time} & \cmark & \xmark & \cmark & \xmark & \xmark & \xmark \\\newline
TEMPTABQA \citep{gupta2023temptabqa} & \cmark & \xmark & \cmark & \xmark & \xmark & \cmark \\\newline
TEMPREASON \citep{tan2023towards} & \cmark & \cmark & \cmark & \xmark & \xmark & \cmark \\\newline
TIQ \citep{jia2024tiq} & \cmark & \xmark & \cmark & \xmark & \xmark & \cmark \\\newline
TempUN \citep{beniwal2024remember} & \cmark & \xmark & \cmark & \xmark & \xmark & \xmark \\\newline
TGQA \citep{xiong2024large} & \cmark & \xmark & \cmark & \xmark & \xmark & \xmark \\
\midrule
\ourbenchmark{} (ours) & \cmark & \cmark & \cmark & \cmark & \cmark & \cmark \\
\bottomrule
\end{tabular}
\end{table}

\Cref{table:benchmark-comparison} provides a comprehensive comparison of \ourbenchmark{} with existing benchmarks across six key dimensions:

\begin{enumerate}
    \item \textbf{Semantics:} whether the benchmark has semantic-type questions.
    \item \textbf{Arithmetic:} whether the benchmark has arithmetic-type questions.
    \item \textbf{Real-world:} whether the benchmark has questions generated from real-world data.
    \item \textbf{Synthetic:} whether the benchmark has questions generated from synthetic data.
    \item \textbf{Hermetic:} whether the benchmark is sealed off from potential LLM training data.
    \item \textbf{Implicit:} whether the benchmark includes implicit questions.
\end{enumerate}

Our analysis reveals that \ourbenchmark{} is unique in incorporating all these question types while effectively mitigating training data leakage. Notably, TEMPREASON \citep{tan2023towards} only covers one category of the arithmetic operations as defined in Section \ref{sec:arithmetic}.

\section{Description of graph generators.}\label{asec:graph-generators}
\begin{figure}
    \centering
    \includegraphics[width=\textwidth]{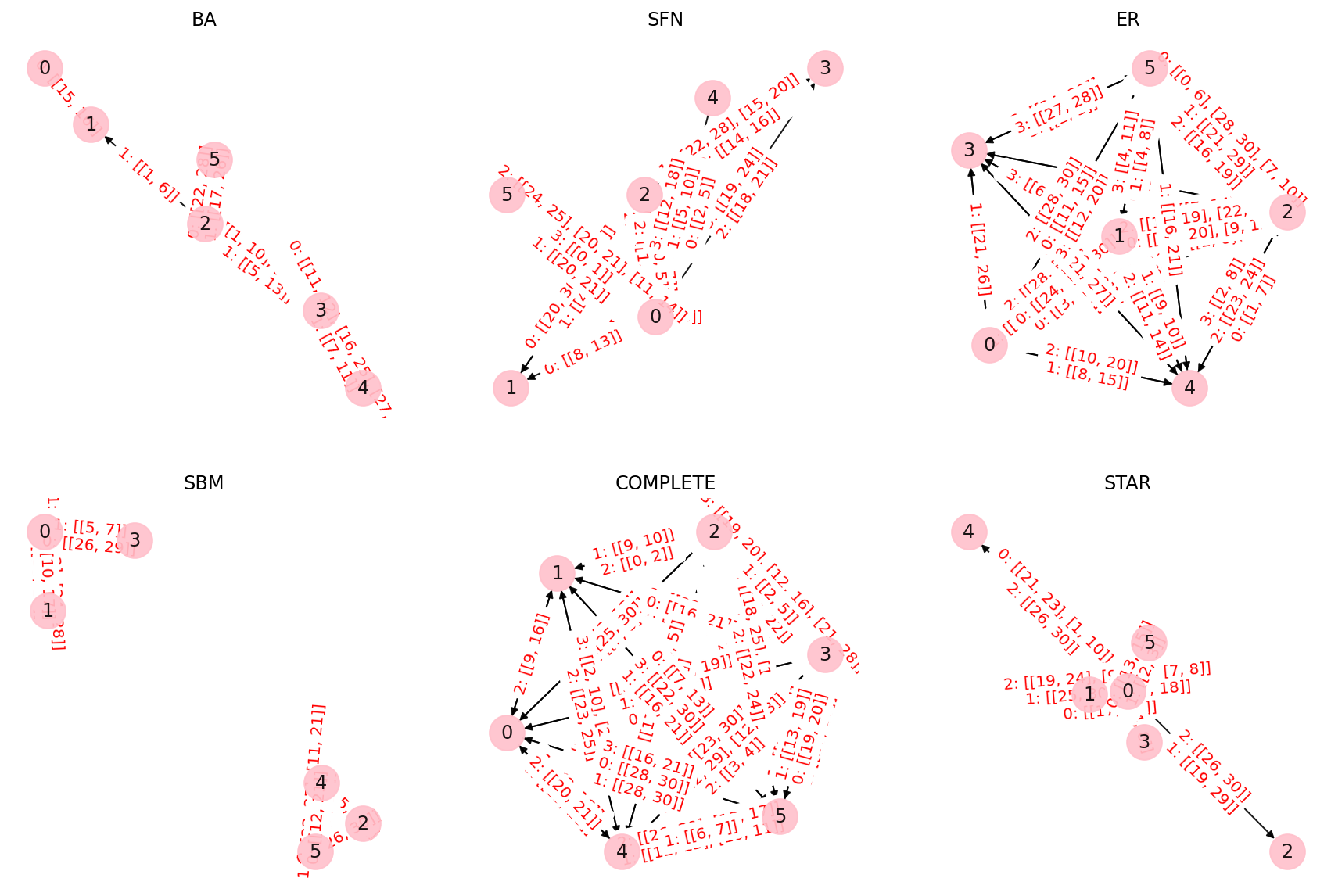}
    \caption{A visualization of a representative graph from each graph generator: Erd\H{o}s-R\'enyi~(ER), Scale-Free Networks~(SFN), Barabási–Albert~(BA), Stochastic Block Model~(SBM), star-graph, and complete-graph.}
    \label{fig:graph-generators}
\end{figure}

Here we detail each graph generator used to create the examples in \ourbenchmark{}. First, we cover the six \emph{random} graph generators used to create the synthetic examples. All random graph generators are probabalistic models which take hyperparameters that control the expected macro-properties of each graph \citep{palowitch2022graphworld}:
\begin{itemize}
    \item Erd\H{o}s-R\'enyi~(ER)~\citep{erdds1959random}: This model takes an edge probability parameter $p$ and generates each edge with probability $p$, i.i.d. over all possible edges.
    \item Scale-Free Networks~(SFN)~\citep{scalefree}: a graph is grown by a sequence of steps, each step either (1) adding a new node and connecting it to an existing node, or (2) adding an edge between two existing nodes. Input parameters control the probability of these events. This process generates a ``scale-free'' power law of node degrees, in sharp contrast to the ER model.
    \item Barabási–Albert~(BA) model~\citep{barabasi}: a graph is grown by a sequence of steps, each step adding a new node to the graph, and connecting the node to $m$ existing nodes with probability proportional to their current degree. Similar to SFN, this process also generates a ``scale-free'' graph with a particular distribution known as the Barabási–Albert model.
    \item Stochastic Block Model~(SBM)~\citep{holland1983stochastic}: This graph model can be thought of as clustered ER. It divides $n$ nodes into $k$ clusters, and then connects two nodes with probability $p$ if they are in the same cluster, else with probability $q$ if they are in different clusters. $k$, $p$, and $q$ are all hyperparameters.
    \item A star-graph generator creates a ``star'' graph on $n$ nodes: node 0 is the center of the star, and all other nodes connect to it (and only it).
    \item A complete-graph generate creates a ``complete'' graph on $n$ nodes, in which all nodes are connect to each other node.
\end{itemize}
An example from each of the above graph generators is shown in \Cref{fig:graph-generators}. In the figure, edges are annotated with temporal relationships in the format \texttt{relation\_id: [interval\_1, ..., interval\_k]}. Note that each edge can have multiple relationships, and each relationship can have multiple intervals. The visualization shows the diversity of temporal knowledge graphs that our framework is able to generate.
We note that while our study was limited to parametric graph generators in this work, the field of graph machine learning \cite{chami2022machine} offers many options for both modeling \cite{perozzi2014deepwalk} and learning \cite{halcrow2020grale,fatemi2021slaps,rozemberczki2021pathfinder,fatemi2023ugsl} link structure.

Second, we describe our Anonymized Wikidata Extract (AWE) strategy for creating anonymized questions from real-world data. We first identify the 78 most common relations in WikiData that specify time-bound entity relationships. Each relation encodes a temporal edge between two entities. To match the schema of our synthetic graphs, we convert each time specification on each edge to an interval. Then, for each entity in the graph, we extract the \emph{ego-graph} of the entity by (1) collecting the entity and all its neighbors and (2) collecting all edges (along with temporal information) between nodes collected in (1). This process produces a temporal graph with a schema identical to those produced from random graph generators. Before generating questions from the graphs, we anonymize them by (a) mapping each entity name to a unique identifier such as \texttt{E679}; and then (b) mapping each relation name to A unique identifier such as \texttt{R3}. We then generate questions from the graph as described in \ref{sec:task-creation}.


\section{Quality of instructions}
A critical question is whether the instructions in the prompts of \arithmeticd{} regarding the format of the output are of high quality and easily understood by large language models (LLMs). Notably, Gemini 1.5 Pro, GPT-4, and Claude-3-Sonnet missed the instructions only $0.05\%$, $0.65\%$, and $4.76\%$ of the time, respectively. This low error rate suggests that the instructions are of high quality. Consequently, we were able to establish a benchmark that allows us to focus purely on the temporal reasoning abilities of the models.

\section{Large-Scale \syntheticd{} Experiments}

To facilitate a more comprehensive analysis and enable deeper insights, we expanded our synthetic dataset significantly. This enlarged dataset now encompasses approximately $50,000$ examples, a substantial increase from the previous set of around $3,000$ examples. We anticipate that this expanded resource will prove valuable for future research endeavors that necessitate a larger and more diverse synthetic dataset.
Due to the computational demands of evaluating all LLMs on this large dataset, results are reported solely for Gemini 1.5 Pro.

\paragraph{Impact of Graph Structure on LLM Accuracy.} Our initial experiment with this expanded dataset involved replicating the graph structure analysis. As illustrated in \Cref{stable:graph-structure}, graph structure continues to exert a significant influence on the final accuracy of the LLM, even within this larger dataset.

\begin{table}[h!]
    \centering
    \caption{LLM temporal reasoning by graph structure on the larger set of \syntheticd{}.}\label{stable:graph-structure}
    \resizebox{0.3\textwidth}{!}{%
    \setlength{\tabcolsep}{3pt}
    \begin{tabular}{lc}
    \toprule
        \textbf{Graph Structure} & \textbf{Accuracy (\%)} \\ 
        \midrule
        BA & 70.96 \\ 
        Complete & 51.07 \\
        ER & 61.85 \\
        SBM & 60.32 \\
        SFN & 79.13 \\
        Star & 73.77 \\
       AWE & 88.72 \\
       \midrule
        Average & 69.40 \\
        \bottomrule
    \end{tabular}
    }
\end{table}

\begin{table}[t]
\caption{Impact of graph structure and question type on a larger set of \syntheticd{}.}\label{stable:question-graph}
\resizebox{0.9\textwidth}{!}{%
\setlength{\tabcolsep}{3pt}
\begin{tabular}{l|ccccccc|r}
\toprule
Temporal task & BA & Complete & ER & SBM & SFN & Star & AWE & Average Rank \\
\midrule
EventAtTimeT & 74.46 & 54.22 & 65.54 & 68.07 & 80.84 & 76.75 & 91.93 & 3.57 \\
EventAtWhatTime & 98.19 & 81.69 & 90.72 & 90.48 & 98.31 & 98.43 & 97.95 & \textbf{1.00} \\
\midrule
BeforeAfter & 53.49 & 34.46 & 48.07 & 45.66 & 68.55 & 58.80 & 73.98 & 7.00 \\
EventAtTimeOfAnotherEvent & 76.99 & 52.89 & 62.53 & 65.18 & 84.82 & 85.78 & 90.48 & 3.79 \\
FirstLast & 70.84 & 49.04 & 61.69 & 55.66 & 87.23 & 68.80 & 92.53 & 4.43 \\
NumberOfEventsInTimeInterval & 57.71 & 40.84 & 54.22 & 49.64 & 64.22 & 70.84 & 83.73 & 6.14 \\
RelationDuration & 88.55 & 80.60 & 83.49 & 82.77 & 87.47 & 88.80 & 90.48 & \textbf{2.36} \\
Timeline & 47.47 & 14.82 & 28.55 & 25.06 & 61.57 & 41.93 & 88.67 & 7.71 \\
\bottomrule
\end{tabular}
}
\end{table}

\paragraph{Impact of graph structure and temporal task on LLM performance.}
Our second experiment examined the accuracy of the model across various question types and graph generators. The expanded dataset provided sufficient examples per category, enabling more robust results. The results are reported in \Cref{stable:question-graph}. Consistent with our earlier findings, single-fact questions generally outperformed multi-fact questions. Notably, the highest accuracy was observed for EventAtWhatTime in single-fact questions and RelationDuration in multi-fact questions. This alignment with the results from the smaller dataset reinforces their significance and suggests that the smaller dataset serves as a reliable proxy for the larger one.

\begin{table}[b]
\centering
\caption{Impact of graph structure and sorting type on a larger set of \syntheticd{}.}\label{stable:order-graph}
\resizebox{0.8\textwidth}{!}{%
\setlength{\tabcolsep}{3pt}
\begin{tabular}{l|ccccccc|r}
\toprule
Order of facts & BA & Complete & ER & SBM & SFN & Star & AWE & Rank Average \\
\midrule
relation\_and\_start\_time & 73.42 & 52.03 & 64.98 & 61.45 & 81.93 & 74.32 & 90.36 & 2.00 \\
shuffle & 66.72 & 44.65 & 54.74 & 54.14 & 74.17 & 72.74 & 85.02 & 4.71 \\
start\_time\_and\_relation & 67.55 & 46.31 & 57.76 & 55.72 & 77.86 & 72.14 & 88.48 & 4.00\\
start\_time\_and\_target & 68.60 & 46.61 & 58.96 & 55.95 & 78.31 & 70.78 & 88.63 & 3.29\\
target\_and\_start\_time & \textbf{78.54} & \textbf{65.74} & \textbf{72.82} & \textbf{74.32} & \textbf{83.36} & \textbf{78.84} & \textbf{91.11} & \textbf{1.00}\\
\bottomrule
\end{tabular}
}
\end{table}

\paragraph{Impact of Graph Structure and order of facts on LLM Performance.} In this experiment, we evaluated LLM performance across various combinations of graph structure and fact order. The results, presented in Table \Cref{stable:order-graph}, reveal that the target\_and\_start\_time ordering consistently yields the best performance across the expanded dataset, regardless of graph structure. Conversely, the shuffle ordering consistently underperforms across most graph structures.

\section{Instructions to participants}\label{asec:crowd-sourcing}
For the crowd-sourcing section in creating the \arithmeticd{} dataset (Expand step), we gave the following instructions to the annotators.

\begin{tcolorbox}[colback=white, colframe=black, title=Time Arithmetic Benchmark Compilation]

Thank you for participating in our eval hour to help us expand our dataset to cover all the categories of time arithmetic that we can think of.

\paragraph{Terminology:}

\begin{itemize}
    \item \textbf{Time arithmetic:} Calculations with time values, often involving years, months, days, hours, minutes, seconds.
    \item \textbf{Category:} A high-level category of time arithmetic operations, such as addition/subtraction, time conversion, etc.
    \item \textbf{Examples:} Real-life sentences that fall into a category. For instance, "Today is 27 July 2020 and I was told that my furniture will be delivered to me in exactly 60 days from now. On what date will the furniture be delivered?" is an example of addition.
\end{itemize}

\paragraph{Goal:}

Our goal is to cover as many real-life categories and subcategories related to time arithmetic as possible. We also want each subcategory to have multiple different real-life examples.

\paragraph{Levels of Importance of Contributions:}

\begin{enumerate}
    \item Discovering/adding a new category.
    \item Adding new real-life examples within a subcategory (please contribute more in less densely populated areas).
\end{enumerate}

Corner cases are useful, but please don't focus all your time on them. Discovering broader categories would be the most useful!\\

Please try to add new examples which are as different from existing ones as possible.\\

Thanks!
\end{tcolorbox}

\section{Limitation and future work}\label{asec:limitation}
The current work has several limitations that provide avenues for future research:

\paragraph{Single-Sentence Time Anchoring .} This benchmark focuses on scenarios where the start and end times of a fact are both mentioned within a single sentence. However, in real-world scenarios, temporal information can be spread across multiple sentences or even documents.
It is worth noting that this setup is easily convertible to the more general case where temporal information can be spread across multiple sentences. While we chose to focus on the single-sentence setup for this initial work, future research could readily adapt the benchmark to the multi-sentence scenario and explore the challenges and opportunities it presents.

\paragraph{Exclusive Focus on Explicit Temporal Facts (By Design).} This benchmark intentionally focuses solely on explicit temporal facts (those with clear time anchors), excluding static facts (those without time anchors). This deliberate choice was made to ensure the benchmark specifically targets and assesses models' capabilities in temporal reasoning. However, future work could expand the scope to include static facts, offering a more comprehensive evaluation of both temporal and general factual reasoning.

\section{Potential negative societal impacts}\label{asec:neg-impact}
While our research aims to enhance the temporal reasoning capabilities of LLMs, it is important to acknowledge potential negative societal impacts. The advancements made in temporal reasoning could potentially be exploited to generate misleading or manipulative content, such as fake news articles with fabricated timelines. Additionally, if these models are primarily accessible to certain groups or organizations, it could lead to an uneven distribution of information power, potentially exacerbating existing societal inequalities. Furthermore, as LLMs become more adept at understanding and generating temporal information, there could be an increased risk of privacy violations, as these models might inadvertently reveal sensitive personal details or historical events that individuals would prefer to keep private. It is crucial to consider these potential negative consequences alongside the technological advancements to ensure responsible and ethical development of LLM technologies.

\section{Dataset documentation: datasheet}
In this section, we follow the recommendations in \citet{gebru2021datasheets}  to provide comprehensive documentation for our dataset.

\subsection{Motivation}

\begin{itemize}
    \item \textbf{For what purpose was the dataset created?} The dataset was created to address the limitations of existing temporal reasoning benchmarks and to provide a more comprehensive and controlled assessment of LLMs temporal reasoning abilities. The dataset aims to provide a more rigorous and informative evaluation of LLM temporal reasoning abilities, contributing to the development of more advanced and reliable AI systems.
    
    \item \textbf{Who created the dataset (\eg{} which team, research group) and on behalf of which entity (\eg{} company, institution, organization)?}
    The dataset was created by a team of ML researchers at Google all listed as authors of the paper. The authors represent various divisions within Google, including Google Research, Google DeepMind, and Google Cloud teams.

    \item \textbf{Who funded the creation of the dataset?} The dataset was created by Google employees as part of their work at the company, so Google funded its development.

\end{itemize}

\subsection{Composition}

\begin{itemize}
    \item \textbf{What do the instances that comprise the dataset represent (\eg{} documents, photos, people, countries)?}
    Each instance within the dataset represents a text-only data point as sentences describing the reasoning problem to be used as a prompt for the LLM.
    
    \item \textbf{How many instances are there in total (of each type, if appropriate)?} Our \ourbenchmark{} benchmark contains $46,480$ data points in the \syntheticd{} dataset and $2,800$ data points in the \arithmeticd{} dataset. A sample of size $2800$ from \syntheticd{} is also created for smaller scale experiments.
    
    \item \textbf{Does the dataset contain all possible instances or is it a sample (not necessarily random) of instances from a larger set?} We included both larger set and smaller sample set.
    
    \item \textbf{What data does each instance consist of?} Each data point within the dataset represents a text-only data point as sentences describing the reasoning problem to be used as a prompt for the LLM.
    
    \item \textbf{Is there a label or target associated with each instance?} Each data point has a label as ground truth associated to that.
    
    \item \textbf{Is any information missing from individual instances?} No. All instances have both their text and label available.
    
    \item \textbf{Are relationships between individual instances made explicit (\eg{} users’ movie ratings, social network links)?} There is no relationship between different instances.
    
    \item \textbf{Are there recommended data splits (\eg{} training, development/validation, testing)?} This dataset serves as a benchmark for evaluating LLMs on temporal reasoning and therefore only a test split has been provided to evaluate models on it.
    
    \item \textbf{Are there any errors, sources of noise, or redundancies in the dataset?} The dataset creation process is automated, but manual sampling indicates no errors were found in the examined subset. While rare errors may still exist, they haven't been detected in the samples reviewed.
    
    \item \textbf{Is the dataset self-contained, or does it link to or otherwise rely on external resources (e.g., websites, tweets, other datasets)?} The dataset is self-contained and does not link to external resources.
    
    \item \textbf{Does the dataset contain data that might be considered confidential (e.g., data that is protected by legal privilege or by doctor–patient confidentiality, data that includes the content of individuals’ non-public communications)?} The dataset does not contain any confidential or sensitive information.
    
    \item \textbf{Does the dataset contain data that, if viewed directly, might be offensive, insulting, threatening, or might otherwise cause anxiety?} The dataset does not contain any data that could be considered offensive, insulting, threatening, or anxiety-inducing.
    
    \item \textbf{Does the dataset identify any subpopulations (e.g., by age, gender)?} The dataset does not identify or contain any information that would allow for the identification of subpopulations based on attributes.
    
    \item \textbf{Is it possible to identify individuals (i.e., one or more natural persons), either directly or indirectly (i.e., in combination with other data) from the dataset?} The dataset does not contain any information that could be used to directly or indirectly identify individuals, either on its own or in combination with other data.
    
    \item \textbf{Does the dataset contain data that might be considered sensitive in any way (e.g., data that reveals race or ethnic origins, sexual orientations, religious beliefs, political opinions or union memberships, or locations; financial or health data; biometric or genetic data; forms of government identification, such as social security numbers; criminal history)?} The dataset does not contain any sensitive data that could reveal attributes.
    
\end{itemize}

\subsection{Collection}

\begin{itemize}
    \item \textbf{How was the data associated with each instance acquired?} Each data point is a temporal reasoning question and is generated automatically using codes. The details can be found in the main paper.
    
    \item \textbf{What mechanisms or procedures were used to collect the data (e.g., hardware apparatuses or sensors, manual human curation, software programs, software APIs)?} The data collection process primarily involved a combination of automated procedures and manual human input.
    The majority of the data collection was performed using software programs and scripts that were developed and executed by the authors. These programs included algorithms and techniques designed to generate and curate the specific types of data required for the dataset (details in the paper).
    The authors played a direct role in curating the seed questions used to initiate or guide certain data collection processes. This human input involved selecting relevant topics or prompts to ensure the collected data met the desired criteria. This combination of automated and manual methods allows for both efficiency and a degree of human oversight, which can help ensure the quality and relevance of the collected data.
    
    \item \textbf{If the dataset is a sample from a larger set, what was the sampling strategy (e.g., deterministic, probabilistic with specific sampling probabilities)?} While the primary dataset isn't derived from another, a smaller subset was created for specific experimental needs. This subset was generated through random sampling, utilizing a fixed seed to ensure reproducibility and allow for consistent comparisons across different runs.
    
    \item \textbf{Who was involved in the data collection process (e.g., students, crowdworkers, contractors) and how were they compensated (e.g., how much were crowdworkers paid)?} The data collection process was conducted solely by the authors of the paper, who are full-time Google employees. As such, they were compensated through their regular salaries and did not receive any additional payment specifically for this task.
    
    \item \textbf{Over what timeframe was the data collected?} The data collection process took place over a four-month period, spanning from March to June, leading up to the project deadline.
    
    \item \textbf{Were any ethical review processes conducted (e.g., by an institutional review board)?} No.
    
    \item \textbf{Did you collect the data from the individuals in question directly, or obtain it via third parties or other sources (e.g., websites)?} While the majority of the dataset was not collected, a small portion of the data collection process did involve direct interaction and collection. 
    
    \item \textbf{Were the individuals in question notified about the data collection?} Yes. The description is provided in the appendix  
    
    \item \textbf{Did the individuals in question consent to the collection and use of their data?} Yes.
    
    \item \textbf{If consent was obtained, were the consenting individuals provided with a mechanism to revoke their consent in the future or for certain uses?} No. The full description is provided in the main appendix of the paper.
    
    \item \textbf{Has an analysis of the potential impact of the dataset and its use on data subjects (e.g., a data protection impact analysis) been conducted?} No.
    
\end{itemize}

\subsection{Uses}

\begin{itemize}
    \item \textbf{Has the dataset been used for any tasks already?} The dataset is used for temporal reasoning in this paper. The scope is temporal reasoning but different models can be evaluated by the same setup.
    
    \item \textbf{Is there a repository that links to any or all papers or systems that use the dataset?} The dataset is generated and used in this paper only.
    
    \item \textbf{What (other) tasks could the dataset be used for?} These datasets can be used for any tasks related to temporal reasoning.
    
    \item \textbf{Is there anything about the composition of the dataset or the way it was collected and preprocessed/cleaned/labeled that might impact future uses?} No.

    \item \textbf{Are there tasks for which the dataset should not be used?} No.
    
\end{itemize}

\subsection{Distribution}

\begin{itemize}
    \item \textbf{Will the dataset be distributed to third parties outside of the entity (e.g., company, institution, organization) on behalf of which the dataset was created?} Yes, the dataset is available publicly in Huggingface.
    
    \item \textbf{How will the dataset will be distributed (e.g., tarball on website, API, GitHub)?}
    The dataset is distributed through Huggingface (\url{https://huggingface.co}), and the code used for generating the dataset will be available in GitHub.
    
    \item \textbf{When will the dataset be distributed?} The dataset is available for reviewers at the time of submission and will be open to public soon.
    
    \item \textbf{Will the dataset be distributed under a copyright or other intellectual property (IP) license, and/or under applicable terms of use (ToU)?}
    In the spirit of open science and collaboration, we have released the datasets under a Creative Commons Attribution 4.0 International (CC BY 4.0) license. For comprehensive details about the terms of the CC BY 4.0 license, please visit the Creative Commons website: \url{https://creativecommons.org/licenses/by/4.0/}.

    \item \textbf{Have any third parties imposed IP-based or other restrictions on the data associated with the instances?} Please refer to the copyright.
    
    \item \textbf{Do any export controls or other regulatory restrictions apply to the dataset or to individual instances?} Please refer to the copyright.
\end{itemize}

\subsection{Maintenance}

\begin{itemize}
    \item \textbf{Who will be supporting/hosting/maintaining the dataset?} Google (the authors listed here in particular) will support, host, and maintain the dataset.
    
    \item \textbf{How can the owner/curator/manager of the dataset be contacted (e.g., email address)?} The owner (Bahare Fatemi) can be contacted through \texttt{baharef@google.com}.
    
    \item \textbf{Is there an erratum?} No. If errors are found in the future, we will release errata using the same link.
    
    \item \textbf{Will the dataset be updated (e.g., to correct labeling errors, add new instances, delete instances)?} Yes, the datasets will be updated as needed to ensure accuracy.
    
    \item \textbf{If the dataset relates to people, are there applicable limits on the retention of the data associated with the instances (e.g., were the individuals in question told that their data would be retained for a fixed period of time and then deleted)?} N/A.
    
    \item \textbf{Will older versions of the dataset continue to be supported/hosted/maintained?} Yes, older versions of the dataset will continue to be maintained and hosted.
    
    \item \textbf{If others want to extend/augment/build on/contribute to the dataset, is there a mechanism for them to do so?} Yes, others are welcome to extend, augment, build on, or contribute to the dataset. They are encouraged to download the dataset, create their own modified versions, and publish their work on their preferred platform.

\end{itemize}

\section{Accessibility}
The datasets created in this research are available for download at \url{https://huggingface.co/datasets/baharef/ToT}.
In the spirit of open science and collaboration, we have released the datasets under a Creative Commons Attribution 4.0 International (CC BY 4.0) license. 
For comprehensive details about the terms of the CC BY 4.0 license, please visit the Creative Commons website: \url{https://creativecommons.org/licenses/by/4.0/}.

\section{Reproducibility}
In our pursuit of reproducibility, we conducted experiments with three prominent model families: Gemini~\cite{gemini1.5}, Claude~\cite{anthropic2024claude}, and GPT-4~\cite{gpt4}. Specifically, we investigated the capabilities of Gemini 1.5 Pro, Claude's Sonnet model, and GPT-4 Turbo. All of these variants are publicly accessible through various platforms, ensuring transparency and enabling further research. We utilized the Google Cloud Platform as our computational infrastructure to execute these experiments. This choice was motivated by its comprehensive suite of machine learning tools it offers.
\end{document}